\documentclass[conference]{IEEEtran}
\IEEEoverridecommandlockouts
\usepackage{cite}
\usepackage{amsmath,amssymb,amsfonts}
\usepackage{algorithmic}
\usepackage{graphicx}
\usepackage{textcomp}
\usepackage{xcolor}
\usepackage{multirow}
\usepackage{booktabs}
\usepackage{makecell}
\usepackage{makebox}
\usepackage{colortbl}
\definecolor{gray}{rgb}{0.95,0.95,0.95}
\def\BibTeX{{\rm B\kern-.05em{\sc i\kern-.025em b}\kern-.08em
    T\kern-.1667em\lower.7ex\hbox{E}\kern-.125emX}}
\begin{document}

\title{Is Large Language Model Good at Triple Set Prediction? An Empirical Study\\
\thanks{\textsuperscript{*}Corresponding Author.}
}

\author{\IEEEauthorblockN{ Yuan Yuan}
\IEEEauthorblockA{
\textit{School of Software Technology} \\
\textit{Zhejiang University}\\
Ningbo, Zhejiang, China \\
yuanyuanyy@zju.edu.cn}
\and
\IEEEauthorblockN{ Yajing Xu }
\IEEEauthorblockA{
\textit{College of Computer Science of Technology} \\
\textit{Zhejiang University}\\
Hangzhou, Zhejiang, China \\
yajingxu@zju.edu.cn}
\and
\IEEEauthorblockN{ Wen Zhang\textsuperscript{*} }
\IEEEauthorblockA{\textit{School of Software Technology} \\
\textit{Zhejiang University}\\
Ningbo, Zhejiang, China \\
zhang.wen@zju.edu.cn}
\and
% \and
% \and
}

\maketitle

\begin{abstract}
% This document is a model and instructions for \LaTeX.
% This and the IEEEtran.cls file define the components of your paper [title, text, heads, etc.]. *CRITICAL: Do Not Use Symbols, Special Characters, Footnotes, 
% or Math in Paper Title or Abstract.
The core of the Knowledge Graph Completion (KGC) task is to predict and complete the missing relations or nodes in a KG. Common KGC tasks are mostly about inferring unknown elements with one or two elements being known in a triple. 
In comparison, the Triple Set Prediction (TSP) task is a more realistic knowledge graph completion task. It aims to predict all elements of unknown triples based on the information from known triples.
In recent years, large language models (LLMs) have exhibited significant advancements in language comprehension, demonstrating considerable potential for  KGC tasks. However, the potential of LLM on the TSP task has not yet to be investigated. Thus in this paper we proposed a new framework to explore the strengths and limitations of LLM in the TSP task. 
Specifically, the framework consists of LLM-based rule mining and LLM-based triple set prediction. The relation list of KG embedded within rich semantic information is first leveraged to prompt LLM in the generation of rules. This process is both efficient and independent of statistical information, making it easier to mine effective and realistic rules.  
For each subgraph, the specified rule is applied in conjunction with the relevant triples within that subgraph to guide the LLM in predicting the missing triples. Subsequently, the predictions from all subgraphs are consolidated to derive the complete set of predicted triples on KG. 
Finally, the method is evaluated on the relatively complete CFamily dataset. The experimental results indicate that when LLMs are required to adhere to a large amount of factual knowledge to predict missing triples, significant hallucinations occurs, leading to a noticeable decline in performance. To further explore the causes of this phenomenon, this paper presents a comprehensive analysis supported by a detailed case study. The datasets and code for experiments are available at
https://github.com/zjukg/LLM-based-TSP.

% 知识图谱补全任务的核心在于通过现有的知识图谱结构，预测和补全其中缺失的关系或节点。常见的知识图谱补全任务多是在已知三元组中一个或两个元素的情况下来推测未知元素。而三元组集预测任务是一种更符合现实生活的知识图谱补全任务。它是从已知的三元组信息中预测出未知三元组的所有元素。近年来，大型语言模型（LLM）强大的语言理解能力使其在KGC任务中展现出巨大的潜力。而LLM在三元组集预测任务上的潜力还有待研究，因此在本文中我们提出了一个新的框架，来探索LLM在TSP任务中的优势和局限性。具体来说，该框架由基于LLM的规则挖掘和基于LLM的三元组集预测两部分组成。首先利用KG关系列表中丰富的语义信息来提示LLM生成规则。这一生成过程是迅速的且不受到统计信息影响，因而更容易挖掘出有效和符合实际的规则。进一步将KG划分为多个子图，然后对于每一个子图，结合规则与子图中与规则相关的三元组信息来提示LLM预测缺失的三元组，从而得出整个KG上的缺失三元组集。最后，在相对完整的CFamily数据集上评估了该方法。实验结果结果表明在LLM需要遵从给定的大量的事实知识来预测缺失三元组时，会出现显著的幻觉现象，导致其在三个分类指标上的表现显著下降。为深入探讨这一现象的原因，本文进一步结合具体案例进行了详细分析。
% 大型语言模型（LLM）已经取得了明显的优势和广泛的应用，其强大的语言理解能力使其在KGC任务中展现出巨大的潜力。
\end{abstract}


\section{Introduction}
Knowledge Graph (KG)\cite{pan2017exploiting} serves as a framework for representing and storing structured knowledge. It is typically organized in the form of a triple, i.e.,  (head entity, relationship, tail entity). The nodes in KG represent the entities in the triple, while the edges represent the relations between the entities.  In this way, the information is more well-organized and easier to understand, so KG is also widely used in a variety of application scenarios such as search engines\cite{huang2020design}, recommender systems\cite{wong2021improving}, question answering\cite{chen2022lako,thambi2024novel} and so on. 
The construction of large open KGs, such as Freebase\cite{bollacker2008freebase}, DBpedia\cite{bizer2009dbpedia}, Wikidata\cite{vrandevcic2014wikidata}, etc., provides robust knowledge support for various intelligent applications. 

Most KGs are far from complete.
Therefore, the Knowledge Graph Completion (KGC) task has become a prominent research focus in the field of KG. The core of the KGC task is to predict and complete the missing relations or nodes in a KG. Specifically, it can be categorized into head entity prediction $(? ,r,t)$, tail entity prediction $(h,r,?) $, relationship prediction $(h,? ,t)$, instance completion $(h,? ,?) $. Zhang et al.\cite{zhang2024start} suggested that these KGC tasks require knowing at least one or two elements of the missing triples, which does not align with real-world scenarios, and thus a new  KGC task Triple Set Prediction (TSP) is defined and a corresponding method called GPHT is proposed. The goal of the TSP task is to directly predict the head entity, tail entity and relation of missing triples from given triples.

In recent years, Large Language Models (LLMs)\cite{openai2023gpt} have achieved obvious advantages and been widely applied, as well as showing great potential in KGC tasks\cite{yao2023exploring,li2024contextualization,zhang2023making,luo2023chatrule,zhu2024llms}. On the one hand, LLMs have powerful language comprehension and generation capabilities, and can quickly process a large amount of complex textual information to complete the inference of KG. On the other hand, the accurate and interpretable knowledge provided by KG can alleviate the hallucination of LLMs. In related studies, LLMs can be leveraged to generate predicted triples from given inputs\cite{yao2023exploring} or to create valid contexts that can enhance predictive accuracy\cite{li2024contextualization}. Researchers have also merged the structural information of KG into LLMs to realize structure-aware reasoning\cite{zhang2023making}. 
It can be seen that LLM can play an important role in KGC tasks. Therefore, we propose the idea of applying LLM to the task of TSP to analyze the advantages and limitations of LLM in this context. 

Depending on good textual comprehension and generative capabilities, LLM is able to understand different relation names in KG and the connections among them. Without giving triples, LLM can also mine rules based on known relations. Compared to traditional rule mining methods, the process of generating rules utilizing LLM is fast and unaffected by statistical information of the KG. In traditional statistical-based rule mining methods \cite{schoenmackers2010learning,galarraga2013amie,galarraga2015fast}, when the number of a certain type of relationship in a KG is small, the metrics of rules containing this type of relationship will be significantly affected. For example, when there are very few triples in the KG that contain the relation \textit{sisterOf}, the metrics such as confidence and support for rules involving \textit{sisterOf} will significantly decrease. As a result, even though these rules may hold substantial value for specific reasoning tasks, they are often not retained during the rule-mining process.
% , for example, making it difficult for these rules to be retained. 
LLM, on the other hand, mines rules based on the semantics of the relations themselves and the patterns obtained from a large amount of text data during pre-training. Therefore, we believe that LLM is capable of mining richer and more realistic rules, especially when the data in KG is not balanced.
Therefore, we apply LLM to both rule mining and triple set prediction stages and propose the specific methods and experimental ideas as follows.

We first write a list of all relations in KG, including inverse relations, to prompt. Then the LLM is guided by the instructions of the prompt to mine the rules based on the connections between different relations. Eventually, valid rules are selected based on rule quality metrics. 
Further, the KG is divided into subgraphs. The information and rules of the subgraphs are then written into the prompt and provided to the LLM to generate the missing triples. 
Three classification metrics\cite{zhang2024start}, namely Joint Precision ($JPrecision$), Squared Test Recall ($ST Recall$), and TSP score ($F_{TSP}$), are chosen to evaluate the effectiveness of the method.

The final experimental results demonstrated that when LLM needs to comply with a given large amount of factual knowledge to predict the missing triples, significant hallucinations occur, which is mainly manifested by 1)~ the triples used for reasoning do not really exist; and 2)~ the reasoning is not made exactly according to the given rule. We believe that the reason for this phenomenon may be related to the way information is stored in KG and the dependence of LLM on context. 
To enhance the performance of LLM in TSP, future research will focus on optimizing the way LLM acquires relevant triples and handles contextual information, thereby reducing reliance on irrelevant data and mitigating the occurrence of hallucinations.

\section{Related works}
The key to the KGC task is to utilize the information in the KG to speculate and complete the missing relations or nodes in it. Common KGC methods can be categorized into translation-based methods such as TransE\cite{bordes2013translating}, TransH\cite{wang2014knowledge}, RotatE\cite{sun2019rotate}, tensor decomposition-based methods such as DistMult\cite{yang2014embedding}, HolE\cite{nickel2016holographic}, SimplE\cite{kazemi2018simple}, deep learning-based methods such as ConvE\cite{dettmers2018convolutional}, convKB\cite{nguyen2017novel}, CapsE\cite{nguyen2018capsule}. In recent years, Graph Neural Network (GNN) techniques have gained increasing attention in KG complementation tasks. Models such as R-GCN\cite{schlichtkrull2018modeling} and CompGCN\cite{vashishth2019composition} have achieved outstanding performance. The various approaches complement each other and together they drive the research and application of KGC tasks.
With its excellent language comprehension and generation capabilities, LLM shows great potential in KGC tasks as it is able to process large amounts of complex text and extract useful information from it.
Luo et al.\cite{luo2023chatrule} pointed out that existing methods for mining logical rules in KGs face challenges due to high computational costs and a lack of scalability for large-scale KGs. To address these issues, they proposed the ChatRule framework, which utilizes LLMs to generate and optimize logical rules by integrating both the semantic and structural information of KGs, thereby enhancing reasoning performance and interpretability.
Yao et al. \cite{yao2023exploring} proposed a method called KG-LLM. This model utilizes LLMs to generate candidate triples from given text inputs. This approach enables the generation of meaningful knowledge graph completion information even in the absence of structured data.
In response to the ineffective use of structural information in existing methods, Zhang et al. \cite{zhang2023making} proposed the KoPA model. This model integrates structural embeddings into LLMs through pre-training, enhancing the model's reasoning ability.
Zhu et al. \cite{zhu2024llms} conducted a comprehensive evaluation of LLMs in KG construction and reasoning, emphasizing their advantages in reasoning tasks. They further proposed the AutoKG multi-agent approach to further enhance KG construction and reasoning capabilities.
These works have shown that large language models demonstrate significant potential in KGC tasks. Therefore, we focus on applying LLM to the task of triple set prediction, which has not been studied in detail.

% Pan et al. \cite{pan2024unifying} proposed a roadmap for integrating LLMs and KGs, focusing on three complementary frameworks to enhance both technologies.

% \subsection{Maintaining the Integrity of the Specifications}

%%%%%%%%%% 预备知识：
\section{Preliminaries}
\textbf{Background. } %$\mathcal{}$
A KG can be denoted as $\mathcal{G=\left \{ E,R,T \right \}}$, where $\mathcal{E}$ is a set of entities, $\mathcal{R}$ is a set of relations, such as \textit{sisterOf}, and $\mathcal{T}=\{(h,r,t)|h,t\in \mathcal{E},r\in \mathcal{R}\}$ is a set of triples. $h,r$ are called the head entity and the tail entity of a triple. The core of the KGC task is to find those triples that are correct but not in the KG. Specifically, triple classification, relation prediction, and entity prediction are three common KGC tasks, where relation/entity prediction is to predict the missing element with two other elements given.
% 一个KG可以表示为 $\mathcal{G=\left \{ E,R,T \right \}}$，$\mathcal{E}$是实体集合，$\mathcal{R}$ 是关系集合，比如\textit{sisterOf}， 而 $\mathcal{T}=\{(h,r,t)|h,t\in \mathcal{E},r\in \mathcal{R}\}$ 是三元组的集合。$h,r$被称为三元组的头部实体和尾部实体。
% KGC任务的核心在于找出那些正确但不在知识图谱中的三元组。具体来说，三元组分类、关系预测、实体预测是常见的三种KGC任务，其中关系/实体预测是在已知另外两个元素的前提下预测缺失的元素。

\begin{figure*}[!ht]
\centerline{\includegraphics[width=1.0\textwidth]{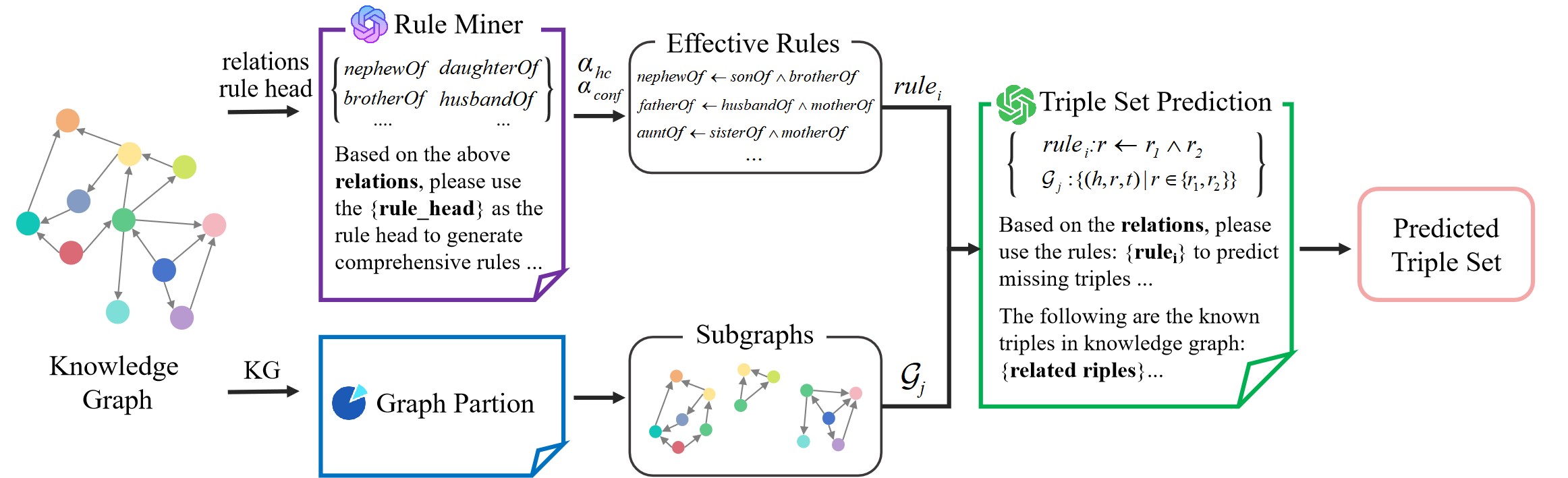}}
\caption{Framework for the TSP task based on Large Language  Models.}
\label{frame}
\end{figure*}

\textbf{TSP Task\cite{zhang2024start} . } %$\mathcal{}$
Given a KG $\mathcal{G}$, the purpose of the TSP task is to discover the missing but correct triple $\mathcal{T}_{predict}$ in the KG. Specifically, the training dataset $\mathcal{G}_{train}=\{\mathcal{E},\mathcal{R},\mathcal{T}\}$ is provided to the model.  After the model predicts the candidate triple, the test set $\mathcal{T}_{test}=\{(h,r,t)|h,t \in \mathcal{E},r\in \mathcal{R},(h,r,t)\notin \mathcal{T} \}$ to evaluate the model. 

% The Triple Set Prediction (TSP)定义如：给定一个KG $\mathcal{G}$，TSP任务的目的在于发现知识图谱中遗漏但正确的三元组$\mathcal{T}_{predict}$. 具体来说，将训练数据集$\mathcal{G}_{train}=\{\mathcal{E},\mathcal{R},\mathcal{T}\}$提供给模型，待模型预测出候选三元组后，利用测试集$\mathcal{T}_{test}=\{(h,r,t)|h,t \in \mathcal{E},r\in \mathcal{R},(h,r,t)\notin \mathcal{T} \}$来评估模型. 

% \textbf{Logical Rule. }一条逻辑规则$l$可以表示为：
% \begin{equation}
% r_{h}(X,Y) \gets r_1(X,Z_1)\wedge  \dots \wedge r_n(Z_{n-1},Y)\label{Rule}
% \end{equation}
% 其中$r_{h}(X,Y)$代表规则头，$r_1(X,Z_1)\wedge \dots \wedge r_L(Z_{L-1},Y)$代表规则体，$L$代表规则的长度。当右侧的规则体成立时，我们可以得出规则头。$r,r_1, \dots ,r_n$对应KG中的关系，$X,Y,Z_1, \dots ,Z_{L-1}$对应KG中的实体。

\textbf{Logical Rule.} A path-based logic rule that is widely studied\cite{agrawal1994fast,zaki1997new,stepanova2018rule,khan2024wisrule,galarraga2013amie} can be expressed as:
\begin{equation}
r_{h}(X,Y) \gets r_1(X,Z_1)\wedge \dots \wedge r_K(Z_{K-1},Y)\label{Rule}
\end{equation}
where $r_{h}(X,Y)$ represents the rule head, $r_1(X,Z_1)\wedge \dots \wedge r_K(Z_{K-1},Y)$ represents the rule body, and $K$ represents the length of the rule. When the rule body holds, we can derive the rule head. In the rule, $\{r,r_1, \dots ,r_K\}$ corresponds to relations in KG, while $\{X,Y,Z_1, \dots ,Z_{K-1}\}$ corresponds to entities in KG. For convenience and brevity, we can abbreviate the rule as $r \gets r_1 \wedge r_2 \wedge \dots \wedge r_K$ without causing confusion.

%%%%%%%%%% 方法
\section{Method}
In this subsection, we present the specific framework and composition of the proposed LLM-based triple set prediction model. The framework of the model is shown in Figure. \ref{frame}, which includes three parts: rule mining based on LLM, graph partitioning, and triple set prediction.

% 在本小节中，我们将介绍所提出利用大语言模型进行三元组集预测的具体框架与组成。该方法的framework如Figure. \ref{frame}所示，其中包括了基于LLM的规则挖掘、图划分、三元组集预测三个部分。
% 我们首先将KG中的关系列表写入prompt，并提供给LLM，通过prompt的指示引导LLM根据不同关系之间的联系来挖掘规则，之后选择出指标大于阈值的有效规则。进一步将知识图谱划分为多个子图。然后将子图的信息与规则写入prompt，并提供给LLM来生成最终预测的缺失三元组。

\subsection{Rule Miner Based on LLMs}
For a given KG $\mathcal{G}$, we first add the inverse triple of each triple to KG, i.e., $\mathcal{G} \gets \mathcal{G}\cup \{(h,r^{-1},t ) | (h,r,t)\in \mathcal{G}\}$. Specifically, we generate the corresponding inverse relation by adding the prefix “inv\_” to each relation, i.e., $\mathcal{R} \gets \mathcal{R}\cup \{r^{-1}| r\in \mathcal{R}\}$. 

\begin{figure}[!htbp]
\centerline{\includegraphics[width=3.5in]{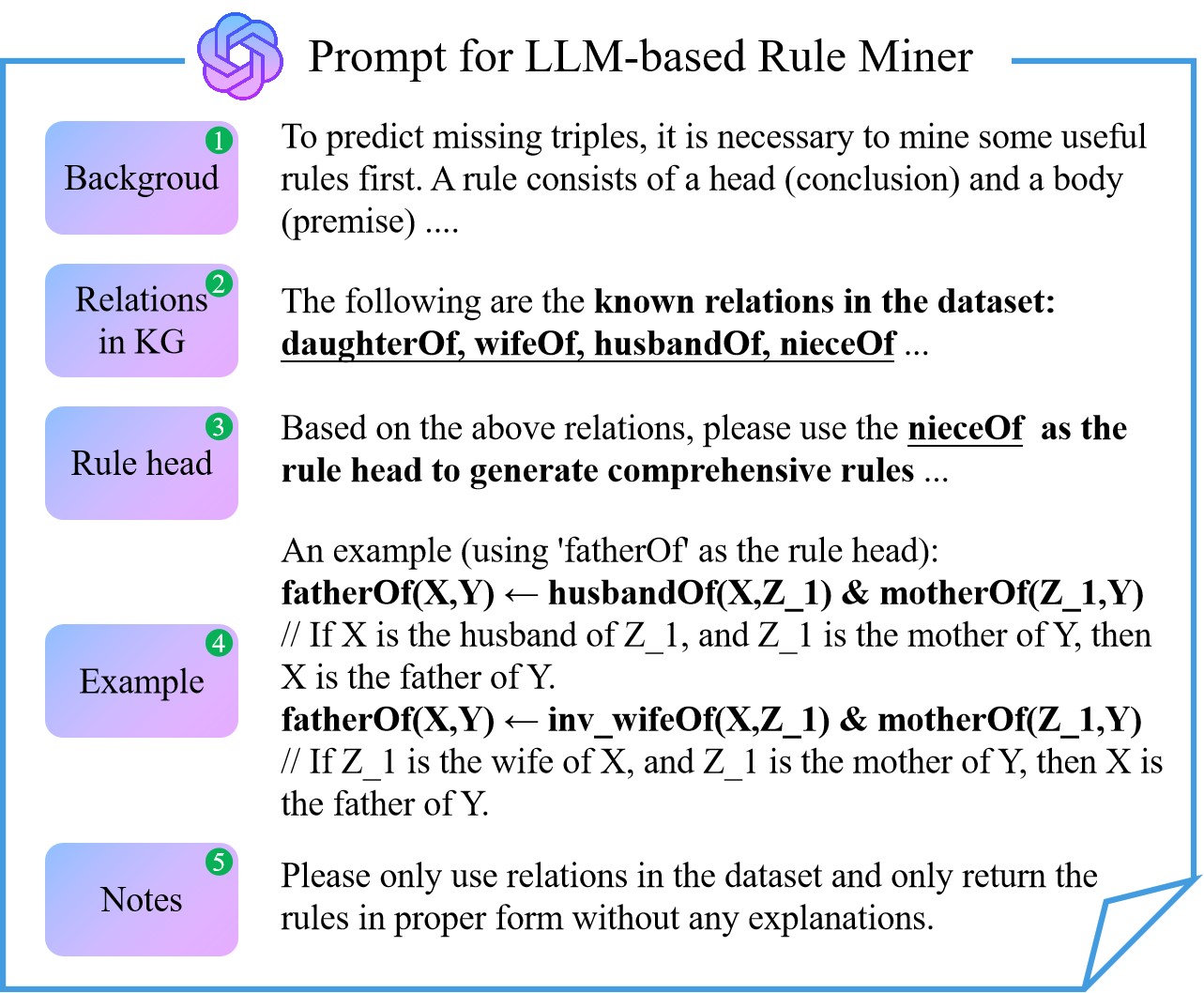}}
\caption{A specific prompt for LLM-based rule miner with \textit{nieceof} as the rule header.}
\label{Prompt4rule}
\end{figure}

Since the large language model has been trained on a large corpus, it is capable of understanding the connections between different relations. Therefore, we utilize the LLM to mine rules based on the relations in the KG. Specifically, we design a detailed prompt which is shown in Fig. \ref{Prompt4rule}. We can see that the prompt consists of five parts: background, relations in KG, rule head, example, and notes. In the background part, we describe the KG as well as the composition and definition of logical rules. Next, we provide all the relations in the KG for LLM. We then specify the rule header as one of the relations in KG (e.g. \textit{nieceOf}) and then direct the LLM to generate rules based on the given lits of relations. To make LLM better understand the problem and answer the questions, we give a concrete example. The example gives two rules with \textit{fatherOf} as the rule head and the corresponding logical explanations. At the end of the prompt, we provided some guidelines to ensure that the LLM generates rules in the correct format. 
% Since the large language model has been trained on a large corpus, it is capable of understanding the connections between different relations. In order to fully utilize the inference ability of the LLMs, we first explicitly list all the relations $\mathcal{R}$ and write them as input data to the prompt. 
Then, we generate rules by setting each relation in turn as the rule head. In this way, LLM analyzes the interconnections and logical relationships among different relations based on the information in prompt and then generates effective rules. Ultimately, we utilize two commonly used metrics \textit{confidence} (conf) and \textit{head coverage} (hc) \cite{lajus2020fast} to filter out high-quality rules from the generated results. Specifically, we retain the rules that confidence and head coverage are higher than the thresholds $\alpha_{conf} = 0.45$ and $\alpha_{hc} = 0.05$.

% 对于给定的KG $\mathcal{G}$，我们首先将其中每个三元组的逆三元组添加到KG中，即，$\mathcal{G} \gets \mathcal{G}\cup \{(h,r^{-1},t ) | (h,r,t)\in \mathcal{G}\}$。具体地说，我们通过对每个关系添加“inv\_”前缀来生成对应的逆关系，即，$\mathcal{R} \gets \mathcal{R}\cup \{r^{-1}| r\in \mathcal{R}\}$。

% 由于大语言模型已经在大量语料库上进行训练，具备了强大的语言理解和生成能力，能够较好地理解不同关系之间的联系。
%%% 更新
% 【new】因此，我们利用LLM根据KG中的关系来挖掘规则。具体来说，我们设计了一个较为详细的prompt如图所示。我们可以看出，prompt共包括五个部分：背景、KG中的关系、规则头、例子和notes. 在背景部分，我们描述了KG以及规则的组成和定义。接着，我们为大模型提供了KG中的所有关系。然后我们指定规则头为关系列表中的某一关系（如nieceOf），然后引导LLM基于所给的关系来生成规则。为了让LLM更好地理解问题和回答问题，我们给出了一个具体的例子。例子给出了以fatherOf作为规则头的两条规则和相应的逻辑性的解释。在prompt的最后，我们还给出了一些提示让LLM生成正确格式的规则。
% 【old】为了充分利用大语言模型的推理能力，我们设计了较为详细的prompt如图所示。% 首先明确地列出所有关系$\mathcal{R}$，并将这些关系作为输入数据写入prompt中。

% 接着，我们通过遍历这些关系，依次将每个关系设定为规则的头部。通过这种方法，LLM会根据prompt所提供的信息逐一分析不同关系之间的相互关联性和逻辑关系，进而生成以每个关系作为规则头的规则。我们利用两种常用的指标\textit{confidence} (conf)和\textit{head coverage} (hc)\cite{lajus2020fast}来从提取的规则中筛选出高质量的规则。Specifically, we retain the rules that confidence and head coverage are greater than the thresholds $\alpha_{conf} = 0.45$ and $\alpha_{hc} = 0.05$.

% prompt的具体设计如【图】所示。
% 规则筛选：

% \subsection{Graph Partition}\label{1}

\subsection{Triple Set Prediction with LLMs}
Considering the fact that the LLM input length is limited, KGs with a large number of triples cannot be input at once, we divide the knowledge graph and then perform triple set prediction for each subgraph individually.
% 考虑到LLM输入长度有限的事实，三元组数量较多的知识图谱无法一次性输入所有三元组，因此我们考虑对知识图谱进行划分，然后对划分出的每个子图分别进行三元组集预测。

\subsubsection{Graph Partition}
% Graph partitioning methods can be categorized into vertex-cut partition and edge-cut partition according to the cut objects. But the two methods have the problems of missing edges and duplicated nodes, respectively.Thus, 
We follow the “soft” vertex-cut KG partition method proposed in \cite{zhang2024start} to obtain subgraphs from KG. 
This method allows subgraphs to share some of their entities with each other, thus ensuring the completeness of information in the division process as much as possible. It consists of two main steps: primary entity grouping and entity group fine-tuning. 
% Its goal is to assign all entities to different entity groups so that the set of ungrouped entities $\mathcal{E}_U$ is eventually empty. 

In the primary entity grouping stage, we first initialize the ungrouped entity set $\mathcal{E}_U=\mathcal{E}$ and the grouped entity set $\mathcal{E}_G=\{ \}$, and divide $\mathcal{E}$ into a number of disjoint entity subsets. The smaller entity sets are then processed and the entity sets with the proper size are removed from $\mathcal{E}_U$ and added to the entity group set $\mathcal{E}_G$. 
Then an entity is selected from $\mathcal{E}_U$ and its neighbor entity set within L-hop is extracted. Finally, the set of neighbor entities within L-hop is added to $\mathcal{E}_G$, and the set of neighboring entities within L-1 hop is removed from $\mathcal{E}_U$. This process continues until all entities in $\mathcal{E}_U$ have been traversed.

In the entity group fine-tuning stage, a random entity $e$ is selected from $\mathcal{E}_U$, then the neighboring entities of $e$ are merged into the smallest entity group $\mathcal{E}^{'}$ that contains $e$, and $e$ is removed from $\mathcal{E}_U$. When all entities are grouped, we obtain the size-balanced entity group set $\mathcal{E}_G$. 
Finally, for each $\mathcal{E}^{'} \in \mathcal{E}_G$, the subgraph $\mathcal{G}_{\mathcal{E}^{'}}$ is constructed by adding triple with head and tail entities in $\mathcal{E}^{'}$, ultimately getting the set of subgraphs $\mathcal{ G}_{part}=\{\mathcal{G}_1, \mathcal{G}_2, ... , \mathcal{G}_k\}$.

% 图分割方法可按照割分对象分为vertex-cut partition and edge-cut partition两种，但两种方法分别存在边缺失和节点重复的问题。Thus, we following the “soft” vertex-cut KG partition method proposed in \cite{zhang2024start} to obtain subgraphs from KG. 该方法使得子图之间可以共享部分实体，从而尽可能地确保了划分过程中信息的完整性。该方法主要包括了primary entity grouping and entity group fine-tuning两个步骤。其目标是将所有实体分配到不同的实体组中，从而最终使得未分组的实体集合$\mathcal{E}_U$为空。 

% 在primary entity grouping阶段，首先初始化未分组实体集$\mathcal{E}_U=\mathcal{E}$，已分组实体集$\mathcal{E}_G=\{ \}$，并将$\mathcal{E}$划分为多个互不相交的实体子集。
% 然后根据设定的最小和最大实体数量阈值，处理较小的实体集，并将大小合适的实体集从$\mathcal{E}_U$移除，并添加到entity group set $\mathcal{E}_G$。然后从$\mathcal{E}_U$选择一个实体，提取其L跳内邻居实体集，其中使用动态概率$p_i$控制邻居实体集的大小。最终，将L跳内邻居实体集添加到$\mathcal{E}_G$，并从$\mathcal{E}_U$中移除L-1跳内邻居实体集，知直到遍历完$\mathcal{E}_U$中的所有实体。

% 在entity group fine-tuning阶段，从$\mathcal{E}_U$中随机选择一个实体 $e$，将$e$的邻居实体合并到包含$e$的最小entity group $\mathcal{E}^{'}$，并从 $\mathcal{E}_U$中移除$e$，迭代至所有实体被分组，从而得到大小平衡的entity group set $\mathcal{E}_G$。最终，对于每个$\mathcal{E}^{'} \in \mathcal{E}_G$，
% 通过添加以$\mathcal{E}^{'}$中实体为头尾实体的三元组来构建子图$\mathcal{G}_{\mathcal{E}^{'}}$,最终得到子图集合$\mathcal{G}_{part}=\{\mathcal{G}_1, \mathcal{G}_2, ..., \mathcal{G}_k\}$.

\subsubsection{LLM-based Triple Set Prediction}
When predicting missing triples, we first need to write the information of subgraphs and related rules into a detailed prompt, and then provide the prompt to LLM. Specifically, the prediction process includes three steps: subgraph information extraction, prompt generation, and LLM reasoning.
% When predicting missing triples, there are three steps: subgraph information extraction, prompt generation, and LLM reasoning.

\textit{Subgraph information extraction}: For each subgraph $\mathcal{G}_i \in \mathcal{G}_{part}$, we first extract all known triples from the subgraphs which will be written into the prompt. This step is crucial since the information of triples is the basis for model inference. 
Meanwhile, considering the possible hallucination phenomenon of LLMs, we further filter out the rule-related triples from the subgraph. Specifically, for a rule $r \gets  r_1 \wedge r_2 \wedge \dots \wedge r_m $, we extract the relations contained in the rule body $\mathcal{R}_{rule}=\{r_1, r_2, \dots, r_m \}$. Then the triples containing any relation in $\mathcal{R}_{rule}$ is further extracted from the subgraph $\mathcal{G}_i$, i.e., $\mathcal{T}_i=\{(h,r,t)|h,t\in \mathcal{E}_i,r\in \mathcal{R}_{rule} \}$. 
By providing $\mathcal{T}_i$ to the LLM instead of all triples in the subgraph, it can help mitigate the risk of the LLM using triples that do not actually exist in the subgraph during reasoning. 

\begin{figure}[!htbp]
\centerline{\includegraphics[width=3.5in]{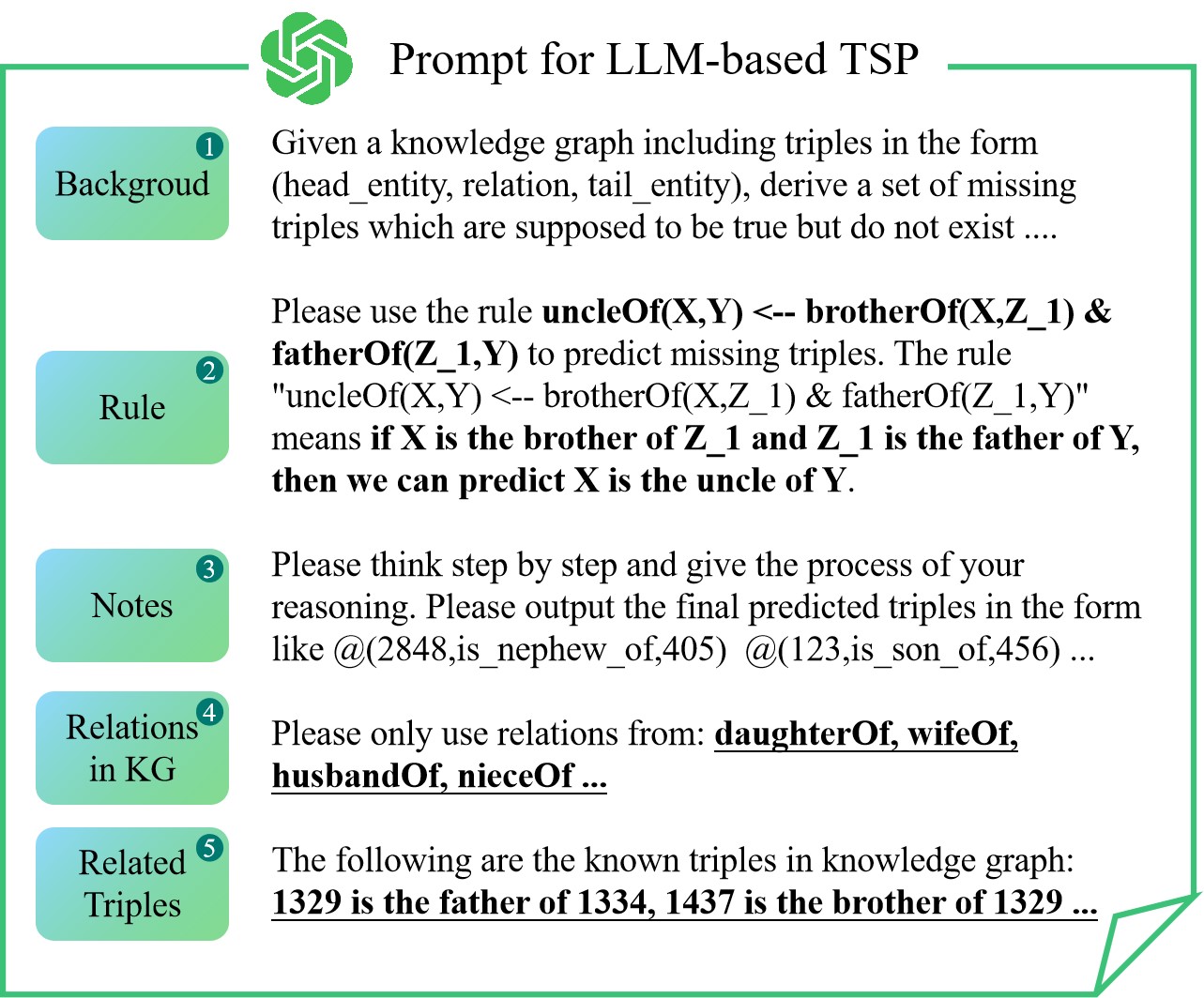}}
\caption{A specific prompt for LLM-based triple set prediction with \textit{$uncleOf(X,Y) \gets brotherOf(X,Z_1) \wedge fatherOf(Z_1,Y)$} as the rule.}
\label{Prompt4tsp}
\end{figure}

\textit{Prompt generation}: We utilize the information of rule-related triples from the subgraph, along with the rule to generate a prompt. We incorporate corresponding explanations for these rules to enhance the prompts. In the large model inference process, we do not provide explicit reasoning examples, which means that this process can be considered a form of zero-shot reasoning. In addition, we introduced the thought of Chain-of-Thought (CoT) \cite{wei2022chain} in the design of Prompt to make LLMs simulate the human's step-by-step thinking process. A specific prompt with \textit{$uncleOf(X,Y) \gets brotherOf(X,Z_1) \wedge fatherOf(Z_1,Y)$} as the rule is shown in Figure \ref{Prompt4tsp}.

\textit{LLM reasoning}: In the reasoning process, LLMs will predict the missing triples based on the information in the prompt. For each subgraph $\mathcal{G}_i \in \mathcal{G}_{part}$, all rules in $\mathcal{R}ule$ will be traversed sequentially to make multiple predictions. After traversing all subgraphs, the predicted triples are extracted from the LLM responses to form the set $\mathcal{T}_{predict}$.

% 在实现三元组集预测任务时，我们首先需要将子图的信息与相关的规则写成详细的prompt，然后将该提示提供给LLM。具体而言，预测过程包括子图信息提取、Prompt生成、LLM推理三个步骤。

% 子图信息提取：对于每个子图$\mathcal{G}_i \in \mathcal{G}_{part}$，我们首先从子图中提取所有已知的三元组，并将这些信息以清晰的方式写入prompt中。这个步骤非常重要，因为这些信息将作为模型推理的基础。同时考虑到大语言模型存在的幻觉现象，我们进一步筛选出子图中与规则相关的三元组。具体来说，对于一条规则$r \gets r \wedge r_1 \wedge r_2 \dots \wedge r_m $，我们提取出规则体中包含的关系类型$\mathcal{R}_{rule}=\{r_1, r_2, \dots, r_m  \}$。进一步从子图$\mathcal{G}_i$中提取出包含$\mathcal{R}_{rule}$中任何关系的三元组，即，$\mathcal{T}_i=\{(h,r,t)|h,t\in \mathcal{E}_i,r\in \mathcal{R}_{rule} \}$。由此能在一定程度上避免大语言模型在推理时使用所生成的并不存在于子图中的三元组。

% Prompt生成：为了使LLM能充分利用KG和规则的信息，我们将子图中的已知信息、适用的规则、KG中的所有关系都写入Prompt中，并为规则添加了相应的解释。在大模型推理过程中，我们并未提供明确的推理示例，因此该过程可以被视为一种零样本推理 。此外，在设计Prompt时，我们引入了CoT的概念来让LLM更好地模拟人类逐步思考的过程来解决问题，从而增强推理结果的准确性。 

% LLM推理：在每一次推理过程中，将输入一个子图和一条规则来生成prompt。模型将基于输入的信息和规则，预测子图中可能缺失的三元组。 对于每个子图$\mathcal{G}_i \in \mathcal{G}_{part}$，会依次遍历$\mathcal{R}ule$中所有规则来进行多次预测。遍历完所有子图后，从LLM的回复中提取出预测的三元组构成三元组集合$\mathcal{T}_{predict}$.

% $\mathcal{R}ule$

%%%%%%%%%% 实验
\section{Experiments}
In this section, we describe the procedure and results of the experiment in detail. We first introduce the dataset, model, and evaluation metrics used in the experiment. Then we analyze the results of our method. Finally, a specific case is used to analyze the significant hallucination phenomenon that occurs when LLM performs the task of triple set prediction.
% 在这一小节中，我们将详细介绍实验的过程和结果。我们首先介绍
% 实验所用的数据集和模型、以及所用的评价指标。进而分析了我们方法在不同指标上的结果。最后用一个具体的case来分析LLM进行三元组集预测任务时所出现的显著的幻觉现象。

\subsection{Settings}
\textbf{Dataset. }
The dataset we use is CFamily, which was proposed in the previous work \cite{zhang2024start}. Specifically, it is a more complete dataset constructed by supplementing missing triples through rules on the basis of an initial set of triples about family. The statistical information of CFamily is shown in Table \ref{Static}.

% \textbf{Dataset. }我们所使用的数据集为CFamily，它是在之前的工作\cite{zhang2024start}中提出的。具体来说，它是在一组关于家庭关系的初始三元组基础上，通过规则补充缺失三元组来构建的一个较完整的数据集，其统计信息如Table \ref{Static}所示。
\begin{table}[!htbp]
\centering 
\caption{Statistical information of CFamily.}\label{Static}
\begin{tabular}{cccclc}
\toprule
\textbf{Dataset} & \textbf{\#Ent} & \textbf{\#Rel} & \textbf{\#Triple} & \multicolumn{1}{c}{\textbf{\#Train}} & \textbf{\#Test} \\
\midrule
CFamily  & 2378  & 12    & 22986    & 18388                       & 4598  \\
\bottomrule
\end{tabular}
\end{table}

\textbf{Model.}  Throughout the experiments, we used GPT-3.5-turbo and GPT-4o as LLMs respectively in both rule generation and triple set prediction phases. In rule generation, we utilize GPT to mine valid rules from KG. The rule generation process is fast and unaffected by statistical information. In the triple set prediction phase, we use the logical reasoning ability of GPT to predict the missing triples from the existing triples. For the parameter settings, we select the rule length $K$ from $\{2,3\}$, $\alpha_{conf} = 0.45$, $\alpha_{hc} =0.05 $.

% \textbf{Model.} 在整个流程中，我们共在规则生成和三元组集预测两个阶段使用了GPT-3.5-turbo作为大语言模型。在规则生成时，我们利用GPT-3.5-turbo对知识图谱信息中不同关系名称的理解能力来挖掘有效的规则。并且规则的生成过程是快速且不会受到KG统计信息的影响。在三元组集预测阶段，我们使用GPT-3.5-turbo的逻辑推理能力，让其根据所提供的规则从已有三元组中预测出缺失的三元组。对于参数设置，we select the rule length $L$ from $\{3,4\}$,  $\alpha_{conf} = 0.45$, $\alpha_{hc} =0.05 $.

% 所具备的基于广泛训练数据集的对上下文信息的理解能力
% 在规则生成和三元组集预测阶段，我们使用GPT-3.5来作为LLM。

\begin{table*}[!htbp]
\centering 
\scriptsize
% \footnotesize
\caption{Rules mined on CFamily by LLMs.}\label{Rules}
% \resizebox{0.85\textwidth}{!}{
% \scalebox{1.1}{
% \resizebox{\linewidth}{!}{
\begin{tabular*}{\linewidth}{c|clccc}
\toprule
\multicolumn{1}{c|}{\textbf{LLMs}}     & \textbf{Index} 
& \multicolumn{1}{c}{\textbf{Rule}}
& \textbf{Support} 
& \textbf{\makecell[c]{Head\\coverage}}
& \textbf{Confidence} \\
\midrule
\multirow{10}{*}{GPT-3.5-turbo} & 1              & $auntOf(X,Y) \gets sisterOf(X,Z_1) \wedge inv\_daughterOf(Z_1,Y)$                         & 364              & 0.15                   & 0.8                 \\
                                & 2                         & \cellcolor{gray}{$auntOf(X,Y) \gets sisterOf(X,Z_1) \wedge motherOf(Z_1,Y) $}                           & 319              & 0.13                   & 0.8                 \\
                                & 3                         & \cellcolor{gray}{$fatherOf(X,Y) \gets husbandOf(X,Z_1) \wedge motherOf(Z_1,Y)$}                         & 184              & 0.21                   & 0.5                 \\
                                & 4                         & \cellcolor{gray}{$fatherOf(X,Y) \gets inv\_wifeOf(X,Z_1) \wedge motherOf(Z_1,Y) $}                      & 196              & 0.22                   & 0.49                \\
                                & 5                         & $fatherOf(X,Y) \gets inv\_wifeOf(X,Z_1) \wedge motherOf(Z_1,Z_2) \wedge sisterOf(Z_2,Y)$       & 145              & 0.16                   & 0.52                \\
                                & 6                         & $nephewOf(X,Y) \gets sonOf(X,Z_1) \wedge brotherOf(Z_1,Y)$                                     & 594              & 0.21                   & 0.77                \\
                                & 7                         & $nieceOf(X,Y) \gets sisterOf(X,Z_1) \wedge nephewOf(Z_1,Y)$                                    & 1427             & 0.63                   & 0.61                \\
                                & 8                         & $nieceOf(X,Y) \gets sisterOf(X,Z_1) \wedge nieceOf(Z_1,Y) $                                    & 1287             & 0.57                   & 0.6                 \\
                                & 9                         & \cellcolor{gray}{$uncleOf(X,Y) \gets brotherOf(X,Z_1) \wedge fatherOf(Z_1,Y)$}                          & 627              & 0.24                   & 0.82                \\
                                & 10                        & $uncleOf(X,Y) \gets brotherOf(X,Z_1) \wedge motherOf(Z_1,Z_2) \wedge sisterOf(Z_2,Y) $         & 262              & 0.1                    & 0.64                \\
\midrule
\multirow{10}{*}{GPT-4o}        & 1                         & $auntOf(X,Y) \gets sisterOf(X,Z_1) \wedge fatherOf(Z_1,Y)$                                     & 532              & 0.21                   & 0.78                \\
                                & 2                         & $auntOf(X,Y) \gets sisterOf(X,Z_1) \wedge inv\_daughterOf(Z_1, Z_2) \wedge inv\_sisterOf(Z_2,Y)$ & 312            & 0.13                   & 0.63                \\
                                & 3                         & $auntOf(X,Y) \gets sisterOf(X,Z_1) \wedge inv\_sonOf(Z_1,Z_2) \wedge inv\_brotherOf(Z_2,Y)$    & 377              & 0.15                   & 0.59                \\
                                & 4                         & \cellcolor{gray}{$auntOf(X,Y) \gets sisterOf(X,Z_1) \wedge motherOf(Z_1,Y)$}                            & 319              & 0.13                   & 0.8                 \\
                                & 5                         & \cellcolor{gray}{$fatherOf(X,Y) \gets husbandOf(X,Z_1) \wedge motherOf(Z_1,Y)$}                         & 184              & 0.21                   & 0.5                 \\
                                & 6                         & \cellcolor{gray}{$fatherOf(X,Y) \gets inv\_wifeOf(X,Z_1) \wedge motherOf(Z_1,Y)$}                       & 196              & 0.22                   & 0.49                \\
                                & 7                         & $nephewOf(X,Y) \gets sonOf(X,Z_1) \wedge inv\_brotherOf(Z_1,Y)$                                & 501              & 0.18                   & 0.78                \\
                                & 8                         & $nephewOf(X,Y) \gets sonOf(X,Z_1) \wedge inv\_sisterOf(Z_1,Y)$                                 & 464              & 0.17                   & 0.79                \\
                                & 9                         & \cellcolor{gray}{$uncleOf(X,Y) \gets brotherOf(X,Z_1) \wedge fatherOf(Z_1,Y) $}                         & 627              & 0.24                   & 0.82                \\
                                & 10                        & $uncleOf(X,Y) \gets brotherOf(X,Z_1) \wedge motherOf(Z_1,Y)$                                   & 364              & 0.14                   & 0.8                 \\ 
\bottomrule
\end{tabular*}
% }
\end{table*}

\begin{table*}[!htbp]
\centering 
\caption{Individual and averaged TSP results for experiments on the CFamily dataset.}\label{TSP results}
% \scalebox{1.1}{
\begin{tabular*}{0.6\linewidth}{@{\extracolsep{\fill}}c|cc|ccc}
% \begin{tabular*}{0.585\textwidth}{c|cc|ccc}
\toprule
\multirow{2}{*}{\textbf{LLMs}} & \multicolumn{2}{c|}{\textbf{\#Triples}}            & \multicolumn{3}{c}{\textbf{Metrics}}             \\ \cmidrule{2 - 6}
                               &$T_{predict}$   & $T_{predict}^+$                   & $JPrecision$ & $ST Recall$ & $F_{TSP}$      \\ 
\midrule
\multirow{4}{*}{GPT-3.5-turbo} & 3583           & 105                               & 0.029       & 0.171       & 0.05        \\
                               & 3287           & 83                                & 0.025       & 0.159       & 0.044       \\
                               & 3338           & 101                               & 0.03        & 0.174       & 0.052       \\\cmidrule{2-6}
                               & 3403±158       & 96±12                             & 0.028±0.3\% & 0.168±0.8\% & 0.049±0.4\% \\ 
\midrule
\multirow{4}{*}{GPT-4o}        & 1444           & 198                               & 0.137       & 0.37        & 0.2         \\
                               & 1169           & 171                               & 0.146       & 0.382       & 0.212       \\
                               & 1216           & 167                               & 0.137       & 0.371       & 0.2         \\\cmidrule{2-6} 
                               & 1276±147       & 179±17                            & 0.14±0.5\%  & 0.374±0.7\% & 0.204±0.7\% \\ 
\bottomrule
\end{tabular*}
% }
\end{table*}

% \subsection{Baseline}

\subsection{Metrics}
Considering that CFamily is a more complete dataset, we will evaluate the effectiveness of our method under the Closed World Assumption (CWA), which assumes that a proposition is considered false if it is not explicitly declared true. Therefore, under the CWA, a triple that does not exist in the KG is considered as a false triple. Based on this assumption, we can categorize the predicted triples into positive and negative triple sets as follows:
% 考虑到CFamily是一个较为完整的数据集，我们将在封闭世界假设(CWA)下对利用LLMs进行三元组集预测的效果进行评估。CWA假定如果某命题未被明确声明为真，则认为其为假。因此，在CWA框架下，知识图谱中未出现的三元组被视为假命题。基于这一假设， 我们可将预测出的三元组分为正三元组集和负三元组集如下：
\begin{align}
\mathcal{T}_{predict}^{+} &= \mathcal{T}_{predict} \cap \mathcal{T}_{test}\label{T+} \\
\mathcal{T}_{predict}^{-} &= \mathcal{T}_{predict} - \mathcal{T}_{predict}^{+}\label{T-}
\end{align}

Following prior work\cite{zhang2024start}, we chose Joint Precision ($JPrecision$), Squared Test Recall ($ST Recall$), and TSP score ($F_{TSP}$) as our three evaluation metrics to evaluate the performance of the proposed method. Under CWA, $JPrecision$ is the percentage of correctly predicted triples in the prediction triples. $ST Recall$ is the percentage of correctly predicted triples in the test set, and the square operation is used considering the large number of triples in the test set. 
As the harmonic mean of $JPrecision$ and $ST Recall$, $F_{TSP}$ effectively balances the two and can accurately reflect the overall performance of the model, especially when the data is imbalanced.
The three metrics under CWA are calculated as shown below:
% Following prior work\cite{zhang2024start}，我们选择了
% Joint Precision ($JPrecision$), Squared Test Recall ($ST Recall$), and TSP score ($F_{TSP}$) 三个的评价指标来衡量模型的表现。在CWA下，$JPrecision$为预测三元组中预测正确的三元组占比，$ST Recall$为测试集中预测正确的三元组占比，并考虑到测试集三元组数量较大而使用了开方运算。而作为$JPrecision$和$ST Recall$的调和平均数，$F_{TSP}$能够很好地平衡二者，并且在数据不均衡时能有效地反映模型的整体表现。在CWA下的三个指标的计算方式如下所示：
\begin{align} %公式最前面加&为左对齐 &&为右对齐
& JPrecision = \frac{ |\mathcal{T}_{predict}^{+}| }{|\mathcal{T}_{predict}|} \label{JP} \\
&STRecall= (\frac{ |\mathcal{T}_{predict}^{+} |}{|\mathcal{T}_{test}|})^{\frac{1}{2} } \label{ST}\\
&F_{TSP}= 2\times \frac{STRecall\times JPrecision  }{STRecall+JPrecision } \label{F}
\end{align}

\subsection{Results}
Table \ref{Rules} shows the statistics of the rules mined by GPT-3.5-turbo and GPT-4o on the CFamily dataset, where the same rules generated by GPT-3.5-turbo and GPT-4o are marked in light gray. It can be seen that these rules are correct and match the relationships between the characters in the family. On the one hand, it can be seen that LLM can mine valid rules based only on the list of relations without giving information about the triples. On the other hand, the number of valid rules mined by LLM at one time is small, thus how to improve the number of valid rules is also a concern for future work.

% Table \ref{Rules}展示了在CFamily数据集上挖掘的规则的统计数据。可以看出这些规则都是正确的，符合家庭中人物之间的关系。一方面，可以看出LLM能够在不给定其余三元组信息的条件下，仅根据关系列表来挖掘出有效的规则。另一方面，LLM一次所挖掘的有效规则数量不多，如何提升有效规则的数量也是后续工作的关注点。

Table \ref{TSP results} shows the individual and average results of experiments conducted three times on the CFamily dataset by using GPT-3.5-turbo and GPT-4o as LLMs. In the three experiments of GPT-3.5-turbo, the average number of predicted triples is 3403, and the average number of correctly predicted triples is 96. The average classification metrics are $JPrecision=0.028,$ $ST Recall=0.168,$ $ F_{TSP}= 0.049$. The fluctuation of each metric is extremely small, 0.3\%, 0.8\%, and 0.4\% respectively. From the experimental results, it can be seen that the proposed model does not perform as well as expected on the test dataset. 
When employing GPT-4o to predict the missing triples, the average number of predicted triples and correctly predicted triples are 1276 and 179 respectively. The average classification metrics are $JPrecision=0.14,$ $ST Recall=0.374,$ $ F_{TSP}= 0.204$. The fluctuation of each metric is also small, 0.5\%, 0.7\%, and 0.7\% respectively. 
It can be observed that employing GPT-4o can enhance the prediction outcome to some degree. This shows that the capability of LLM itself has a significant influence on the TSP task. However, both GPT models demonstrate relatively poor performance on the TSP task. Further analyzing the responses of LLMs, we consider that the poor experimental results may be attributed to the significant hallucination phenomenon generated by LLMs in the prediction process. To further elucidate this phenomenon, we will elaborate on a specific case in the next subsection.

% Table \ref{TSP results}展示了在CFamily数据集上三次实验分别的结果和平均结果。在三次实验中，平均预测的三元组数目为3403，其中平均预测正确的三元组数目为96. 而三次实验的平均分类指标为$JPrecision=0.028,$ $ST Recall=0.168,$ $ F_{TSP}= 0.049$。并且每个指标的浮动并不大，分别为0.3\%, 0.8\%, and 0.4\%. 从实验结果可以看出，我们的方法在测试数据集上的表现不如预期。进一步分析LLM的回复，我们认为实验结果较差的原因可能是LLM在预测过程中产生的显著的幻觉现象。为了进一步阐明这一现象，我们将在下一小节中对具体案例进行详细说明。

% 在准确率、召回率和F1分数上均低于现有基线模型。 

\subsection{Case Study}
\begin{figure}[!htbp]
\centerline{\includegraphics[width=3.7in]{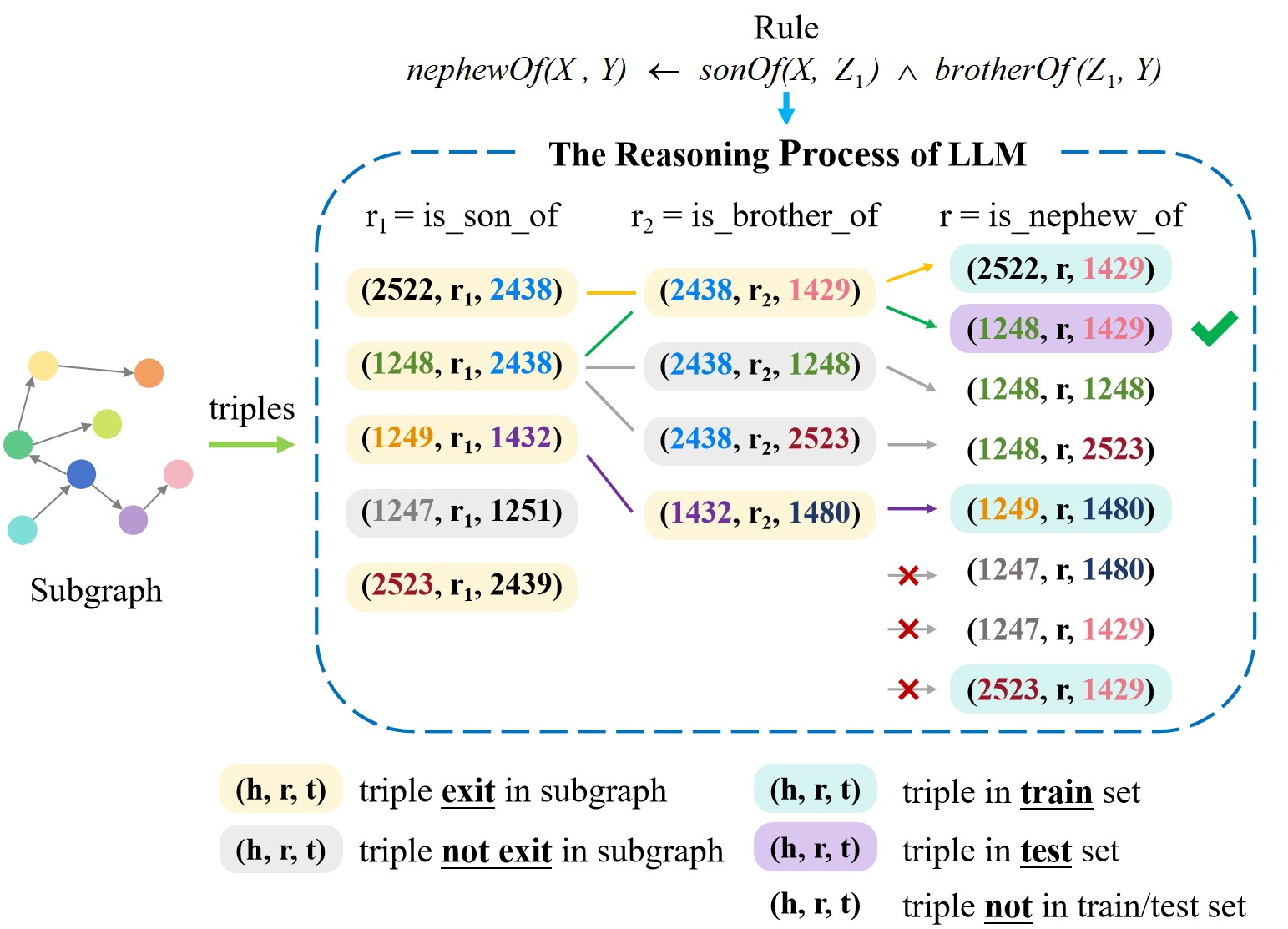}}
\caption{Illustration of the specific case used to demonstrate the hallucination phenomenon observed in the LLM-based TSP task.}
\label{case}
\end{figure}

Fig. \ref{case} shows the specific reasoning process of LLM. As shown in the figure, the triples from the subgraph and a rule are taken as input to the LLM in the form of a prompt. 
Take the rule $nephewOf(X, Y) \gets sonOf(X, Z_1) \wedge brotherOf(Z_1, Y)$ as an example, let $r=is\_nephew\_of, r_1 = is\_son\_of, r_2 = is\_brother\_of$. 
We can observe that during LLM reasoning, one of the  $ r_1$-related triples does not exist in the subgraph, and two of the $ r_2$-related triples do not exist in the subgraph. These non-existent triples are wrong triples generated by LLM, thus showing the very significant hallucination phenomenon of LLM in the reasoning process. This hallucination phenomenon is especially pronounced when we input all the triples of the subgraph into LLM. When only rule-related triples are input, the hallucination phenomenon is mitigated to some extent but still persists. 

In conclusion, we can observe that \textbf{when LLM needs to comply with a given large amount of factual knowledge to predict the missing triples, significant hallucinations occur, which is mainly manifested by 1)~the triples used for reasoning do not really exist; and 2)~the reasoning is not made exactly according to the given rule.} 
% And these hallucinations can lead to low accuracy of prediction results. 
These hallucinations may occur due to the following reasons: 1)~LLM deals with natural language information that is usually unstructured, while the triple in KG is structured information with logical relationships. Therefore LLM may generate hallucination phenomena due to the lack of in-depth understanding of these structured relations. 2)~The entities of the triple in CFamily exist in the form of encodings rather than specific person names in natural language, lacking semantic information in comparison. This may lead to the inability of LLM to accurately understand the roles and relations of the entities. 3)~LLM relies on the current context for reasoning during generation. However, when a large number of facts are involved, the model may not be able to relate the facts correctly, leading to the generation of incorrect information. 

% Fig. \ref{case}展示了一个case中LLM具体的推理过程。如图所示，将子图中的三元组和一条规则作为输入信息，以prompt的形式输入到LLM中。以规则$nephewOf(X, Y) \gets sonOf(X, Z_1) \& brotherOf(Z_1, Y)$为例，设$r=is\_nephew\_of, r_1 = is\_son\_of, r_2 = is\_brother\_of$，我们可知在LLM推理过程中，与$ r_1$有关的三元组中有1个不存在于子图；与$ r_2$有关的三元组中有2个不存在于子图；这些不存在的三元组是LLM生成的错误三元组，由此可见LLM在推理过程中十分显著的幻觉现象。而当我们将整个子图的所有三元组都输入LLM时，这种幻觉现象尤其明显；在最坏的情况下，甚至与其中一个关系有关的三元组都是LLM自己生成的。当只输入与规则相关的三元组时，幻觉现象得到一定程度的缓解，但仍然存在。
% LLM共预测了8个可能的缺失三元组。从推理逻辑的正确性来分析，其中有5个三元组是严格按照规则推理出来的，而其余3个三元组并没有保证$ r_1$所在三元组的尾实体与$ r_2$所在三元组的头实体相同，无法根据规则推理出来。而所预测的三元组中有3个已经存在于训练数据中，只有$(1248, is\_nephew\_of, 1429)$预测正确，存在于测试数据中。

% 综上所述，我们可以发现\textbf{LLM需要遵从给定的大量的事实知识来预测缺失三元组时，会出现非常显著的幻觉现象，主要表现为}\textbf{1)~推理所使用的三元组并不真实存在; 2)~推理时并没有完全按照规则来推理。}而这些幻觉现象会导致预测结果的准确率很低。这些幻觉现象的原因可能有以下几点：1)~LLM处理的自然语言信息通常是非结构化的，而KG中的三元组是具有逻辑关系的结构化信息。因此LLM可能因为缺乏对这些结构化关系的深入理解而产生幻觉现象。2)~CFamily中三元组的实体以编码的形式存在而非自然语言中的具体人名，相比之下缺乏语义信息，这可能导致LLM无法准确地理解实体的角色和关系。3)~LLM在生成内容时依赖于当前上下文进行推理。然而，当涉及到大量事实时，模型可能无法正确地将相关事实联系起来，从而导致错误的信息生成。 

%%%%%%%%%% 结论
\section{Conclusion}
In this paper, we propose a new framework to explore the strengths and limitations of LLM in TSP tasks. Specifically, the KG relation list is first utilized to prompt LLM to mine useful rules. The KG is further divided into subgraphs and the rule-related triples in the subgraphs are extracted. The rule and triples are then utilized to prompt LLM to predict the missing triples. We conducted experiments on the relatively complete CFamily dataset. Experimental results indicate that when a Large Language Model (LLM) needs to follow a substantial amount of factual knowledge to predict missing triples, significant hallucination phenomena occur. We believe that the reason for this phenomenon may be related to the way information is stored in KG and LLM's dependence on context. In the future, we will improve the way of obtaining triples from subgraphs and the reasoning process of LLM to mitigate the illusion phenomenon. Meanwhile, we will further conduct experiments on more datasets.

% 本文中我们提出了一个新的框架，来探索LLM在TSP任务中的优势和局限性。具体来说，首先利用KG关系列表提示LLM挖掘出有用的规则。进一步将KG划分为多个子图，并提取出子图中与规则相关的三元组。然后在每个子图上利用规则和三元组提示LLM预测缺失的三元组。我们在相对完整的CFamily数据集开展了实验。实验结果结果表明在LLM需要遵从给定的大量的事实知识来预测缺失三元组时，会出现显著的幻觉现象。我们认为这一现象的原因可能与KG中信息的存储方式和LLM对上下文的依赖有关。在未来，我们将改进LLM预测时从子图中获取三元组的方式和推理过程，来减轻幻觉现象。同时，我们将进一步在更多数据集上开展实验。

%%%%%%%%%%%% end %%%%%%%%%%%%%%

\section*{Acknowledgment}
This work is founded by National Natural Science Foundation of China (NSFC62306276/NSFCU23B2055/NS FCU19B2027), Zhejiang Provincial Natural Science Foundation of China (No. LQ23F020017), Yongjiang Talent Introduction Programme (2022A-238-G), and Fundamental Research Funds for the Central Universities (226-2023-00138). 

% \section*{References}

\bibliographystyle{IEEEtran}
\bibliography{IEEEabrv,Rreference}
% \bibliography{Rreference}
\end{document}